\documentclass[10pt,journal]{IEEEtran}
\IEEEoverridecommandlockouts
\usepackage{lineno}
\usepackage{fancyhdr}
\usepackage{multirow}
\usepackage{amsmath}
\usepackage{amssymb}
\usepackage{latexsym}
\usepackage{CJK}
\usepackage{subfigure}
\usepackage{indentfirst}
\usepackage{url}
\usepackage{algorithmic}
\usepackage{algorithm}
\usepackage{graphicx}
\usepackage{epstopdf}
\usepackage{longtable}
\usepackage{array}
\usepackage[thmmarks,amsmath]{ntheorem}
\usepackage{diagbox}[2011/11/22]
\usepackage{textcomp,booktabs}
\usepackage[usenames,dvipsnames]{color}
\usepackage{colortbl}
\usepackage{breqn}
\usepackage{subfigure}
\usepackage{flushend}
\usepackage[numbers,sort&compress]{natbib}
\usepackage{color}
\usepackage{engord}

\begin{document}
\title{Task-Oriented Image Transmission for Scene Classification in Unmanned Aerial Systems}
\author{
       Xu Kang,
       Bin Song,~\IEEEmembership{Senior Member,~IEEE},
       Jie Guo,~\IEEEmembership{Member,~IEEE},
       Zhijin Qin,~\IEEEmembership{Senior Member,~IEEE},
       \newline
       and F. Richard Yu,~\IEEEmembership{Fellow,~IEEE}% <-this % stops a space

}
\maketitle

\begin{abstract}
The vigorous developments of Internet of Things make it possible to extend its computing and storage capabilities to computing tasks in the aerial system with collaboration of cloud and edge, especially for artificial intelligence (AI) tasks based on deep learning (DL). Collecting a large amount of image/video data, Unmanned aerial vehicles (UAVs) can only handover intelligent analysis tasks to the back-end mobile edge computing (MEC) server due to their limited storage and computing capabilities. How to efficiently transmit the most correlated information for the AI model is a challenging topic. Inspired by the task-oriented communication in recent years, we propose a new aerial image transmission paradigm for the scene classification task. A lightweight model is developed on the front-end UAV for semantic blocks transmission with perception of images and channel conditions. In order to achieve the tradeoff between transmission latency and classification accuracy, deep reinforcement learning (DRL) is used to explore the semantic blocks which have the best contribution to the back-end classifier under various channel conditions. Experimental results show that the proposed method can significantly improve classification accuracy compared to the fixed transmission strategy and traditional content perception methods.
\end{abstract}
\begin{IEEEkeywords}
Aerial image transmission, task-oriented communication, compressive sensing, reinforcement learning (RL).
\end{IEEEkeywords}

%\enlargethispage*{7mm}

\section{Introduction}
\label{Introduction}
With the exponential growth of the number of mobile edge devices and the huge improvement of computing capabilities of edge devices, increasingly computing demand has been decentralized from the cloud to the edge \cite{1, 2}. Under the coverage of mobile networks, UAVs can provide  services for surrounding users through communication with base stations (BS), thereby creating an environment with high performance, low latency and high bandwidth \cite{3}. Image transmission is one of the most essential tasks that drones need to perform, which can capture high-resolution (HR) images that are unreachable for humans beings in high-altitude environments. Due to the changes in flight area and unstable channel conditions, there are still many challenges for unmanned aerial vehicles (UAV) to maintain low latency in high-definition image transmission scenarios. Therefore, it is necessary to study the transmission strategy and its reliability based on aerial images.

Intelligent image analysis technologies such as image classification, image segmentation, and object detection assist in solving the problems in drone scenes \cite{4}. A key to the artificial intelligence (AI) aided edge computing is how to efficiently implement the AI algorithms with the leverage of cloud and edge \cite{5}. Particularly, if AI algorithms are deployed on front-end drones, high-performance computing cannot be guaranteed due to the limited storage space, computing power, and energy consumption. A task-oriented communication problem arises that the received bits convey the intention of the transmitter. Proper analysis of the semantics conveyed by the target task can guide the transmitter and receiver to achieve a common goal whenever communication occurs \cite{6, 7}. A video coding for machines (VCM) framework for machine vision has been proposed in \cite{8}, where the image content positively contributing to model inference should be preferentially extracted to ensure that the accurate completion of AI tasks.

\begin{figure*}[!h]
	\centering
	\includegraphics[height=57.1443mm,width=176.217mm]{{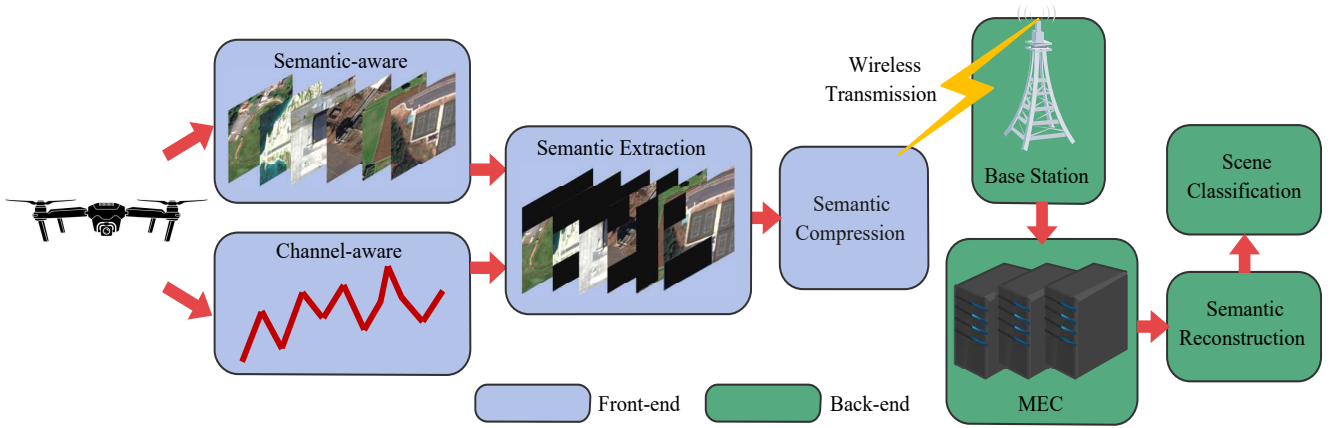}}
	\caption {Illustration of the proposed aerial image transmission framework.}
	\vspace{0cm}
	\label{FIG1}
\end{figure*}

Existing MEC-based image transmission framework tends to either encode the images in terms of the change of channel conditions or compress the images on the basis of their content. However, these two schemes have their own deficiencies. Classical methods such as joint source-channel coding (JSCC) \cite{9, 10} directly maps the image pixel values to the complex-valued channel input symbols, and jointly learns the encoding and decoding neural networks with feedback from channels. With particular focus on the quality of the reconstructed images, they did not consider the entire system as a whole to transmit the image semantics for the back-end machine vision tasks. Some studies extract semantic concepts from the texture structure in images to improve the semantic-preserving ability in down-stream vision tasks \cite{11, 12}. Images are compressed by the global similarity within the context pixels in \cite{13} or the block-based compressive sensing (CS) with different sampling rates based on saliencies of semantic blocks \cite{14}. Nevertheless, these methods only encode the artifacts into bits, which are not tightly coupled with downstream tasks. It is critical to find out useful image pixels with the analysis of transmission tasks. After all, the machine vision system based on convolutional neural network (CNN) understands image concepts differently from humans. 

Inspired by the idea of task-oriented communication, this article aims at building a new aerial image transmission paradigm, as illustrated in Fig. \ref{FIG1}. We consider a remote sensing scenario in which the HR images captured by onboard cameras should be classified immediately. However, the computation resources of the drones are much less than that of the MEC server. There should be a tradeoff between the classification accuracy and transmission latency, which are two main factors determining system performance.

The contributions of this article are summarized as follows:
\begin{itemize}
\item We propose a new aerial image transmission paradigm for the scene classification task. In the training stage, deep reinforcement learning (DRL) is used to explore the semantic blocks which have the best contribution to the back-end classifier under various channel conditions. During the testing stage, the model can independently compress the most essential blocks in the HR images for transmission.

\item Different from traditional task-oriented communications, the proposed algorithm can perceive channel environment by considering the channel gain as the state, which makes it possible to extract semantic blocks for compression under any channel condition. Moreover, the proposed algorithm is suitable for any target model due to the invariant state space and action space.

\item A good balance between transmission latency and classification accuracy is achieved by introducing both of them in the reward function. Experiments show that the system has the ability of making decisions under different channel conditions. Under the premise of transmitting the same amount of bits, the system performance is better than manual designed and traditional content perception methods.
\end{itemize}

The remainder of this paper is organized as follows. Related works are discussed in Section II. Section III gives a detailed description of the system model. Afterwards, the DRL-based system optimization algorithm is proposed in Section IV. Section V verifies the performance of the proposed algorithm  through multiple simulations. Finally, Section VI concludes the paper.

\section{Related Works}
\label{section2}
In this section, we first review the related works on task-oriented communications. We then introduce the state-of-the-art in computer vision with deep reinforcement learning.

\subsection{Task-oriented Communication}
In the era of Internet of Everything, communication plays an important role in the collaborative computing tasks of multiple sensors and computing resources. It is necessary to study the application of task-oriented communication in the future IoT paradigm, which dynamically adjusts the information transmission strategy adapted to the task objective and the underlying changes of the communication environments. The works in \cite{6,7} started from Shannon and Weaver's original definition and categorization of communication, discussing the value of information while proposing the vision of task-oriented communication and its effectiveness from the perspectives of information, communication, control theory, and computer science.

With the rapid improvement of hardware performance, more and more deep learning models are applied in the communication scenarios, even in the task-oriented communication systems \cite{15,16,17,18,19,20}. Xie \emph{et al.} proposed the deep semantic communication (DeepSC) method for wireless text transmission \cite{16}, and its variations, L-DeepSC \cite{15}, DeepSC-S \cite{17}, and MU-DeepSC \cite{19} for complexity reduction, speech transmission, and multi-user transmission with multi-modal data. Particularly, L-DeepSC \cite{15} extends DeepSC to a lite mode, which is affordable for power-constrained IoT devices. \cite{17} is designed for speech signal transmission by realizing semantic information exchange through an attention mechanism. Moreover, \cite{20} extends DeepSC from point to point transmission to a multi-user transmission system, in which the source encoder and channel encoder are both composed of deep networks. Particularly, L-DeepSC \cite{15} extends DeepSC to a lite mode, which is affordable for power-constrained IoT devices. \cite{19} is designed for speech signal transmission by realizing semantic information exchange through an attention mechanism. Moreover, \cite{20} extends DeepSC from point to point transmission to a multi-user transmission system for supporting multimodal data transmission. \cite{17} subsequently introduced Reed Solomon (RS) channel coding to enhance the reliability of transmission. Similarly, an adaptive circular transformer is applied to semantic communication for more flexible transmission \cite{18}.

\subsection{Computer Vision with Deep Reinforcement Learning}
Reinforcement learning (RL) is a classic machine learning method which constructs a Markov decision process (MDP) where agents acquire rewards from environment to update the model parameters in a trial-and-error manner. Deep reinforcement learning (DRL) combines the perception ability of deep learning with the decision-making ability of reinforcement learning. In the high-dimensional continuous space of the state-action pair space, value-based learning will encounter difficulties, and the strategy-based reinforcement learning shows better convergence \cite{21, 22}.

A growing number of researchers are applying DRL to solve the perception-decision problem in computer vision tasks \cite{23}. \cite{24} applies a policy gradient based on random curiosity-driven exploration to image coding, resulting in good performance in Atari games. Uzkent \emph {et al.} select HR image blocks with policy gradients to improve classification accuracy with LR images as states \cite{25}. Yuan \emph {et al.} control the compression of the target model with agent exploring the relationship between the input and output \cite{26}. Hu \emph{et al}. jointly optimize the offloading decision and video coding parameters of the MEC-based face recognition framework with Q-learning \cite{27}. Li \emph{et al}. improve the bit rate by selecting quantization parameters (QPs) for different coding units (CUs) with policy gradients \cite{28}. Li \emph{et al.} propose an image cropping model based on weakly-supervised DRL in \cite{29}. In the same way, Zhang \emph{et al.} construct two agents to move and zoom the bounding-box with the emotional and background information in the image \cite{30}. By simulating an agent as a brush, Zhang \emph{et al}. sequentially relight portraits by DRL and learned a coarse-to-fine policy with local interpretability \cite{31}. Yarats \emph{et al} directly construct a value function in the pixel domain and regularize the learned policy with model-free DRL, so that the image has better features and visual effects \cite{32}.

\section{System Model}

\label{section3}
\begin{figure*}[!h]
	\centering
	\includegraphics[height=82.7326 mm,width=170.6189 mm]{{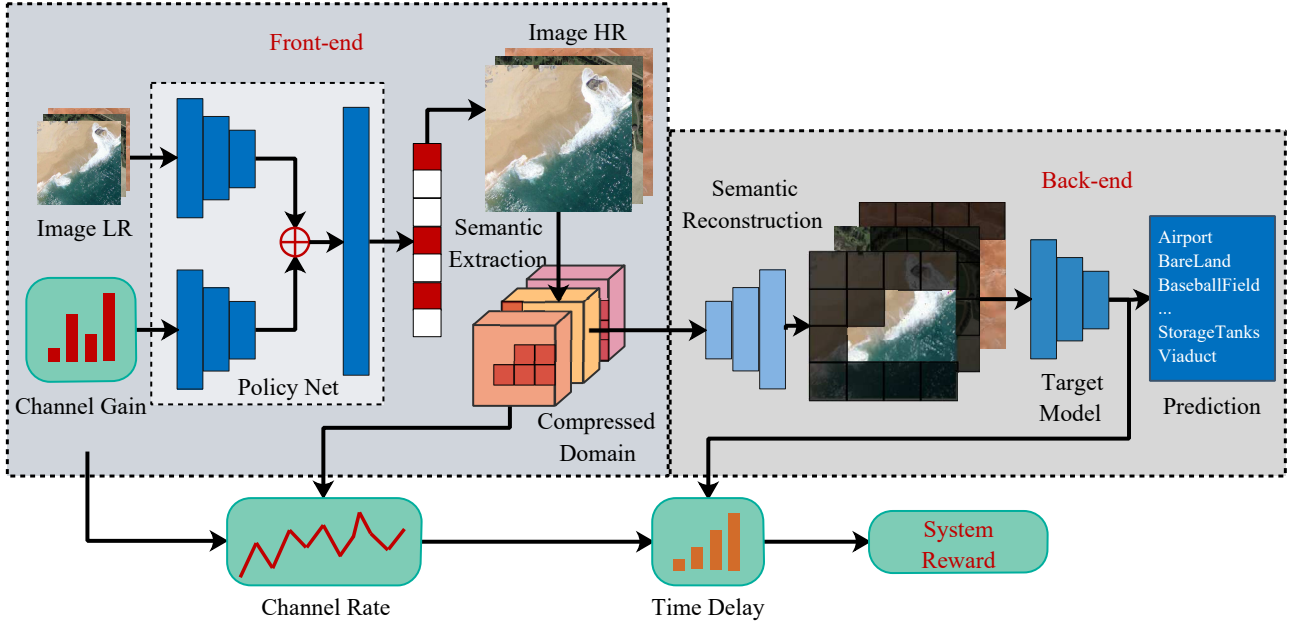}}
	\caption {Proposed task-oriented paradigm for aerial image transmission.}
	\vspace{0cm}
	\label{FIG2}
\end{figure*}

In this section, we propose a task-oriented aerial image transmission framework, as shown in Fig. \ref{FIG2}. The whole system is mainly composed of the front-end UAV and the back-end MEC server, where a semantic extraction network and a block-wise deep compressive sampling model are mainly deployed on the front-end UAV while a reconstruction network and the target model are grouped at the back-end MEC server. Assuming that the structure of the target scene classification classifier is agnostic, we can only optimize and adjust the entire system through the output of the target model, so that the proposed scheme can flexibly adapt to a variety of target models.  With the perception of the channel conditions and the content in LR images before CS, the policy network selects reliable image regions for sampling. When compressed data is transmitted over a wireless channel, the latency is determined by the amount of data and the channel rate. Afterwards, the reconstruction network recovers the image blocks compressed by the semantic extraction network from the compressed data. The reward is calculated by the model prediction and the latency, so as to optimize the whole system. The detail will be introduced in the following.

\subsection{Semantic Extraction Module}
Semantic blocks are defined as a collection of image blocks that contribute the most to the correct classification of the back-end model. The semantic extraction module is to extract effective semantic blocks for compression through the perception of image content and channel gain, which mainly includes a policy network. Due to the limited computing resources on the sensor, directly using HR images for semantic extraction will cause high energy consumption and latency. Aided by the principle of CS sampling, the front-end sensor can capture LR images in advance, which can be regarded as a prerequisite for semantic extraction. As a branch of joint perception, LR images can also provide semantic information about HR images to guide the model to find out critical pixel regions. The other channel sensing branches are able to provide information for channel gains, enabling the model to select semantic blocks at different scales. Then, the output policy network for semantic extraction is defined as follows:
\begin{equation} \label{eq1}
	\textbf{p}=sigmoid(f_{\psi}(f_{\theta}(\textbf{X}^{lr})+f_{\phi}(g)),
\end{equation}
where $\textbf{X}^{lr},g$ denotes the input LR image and channel gain respectively. $f_{\theta}$ is a ResNet18 for image feature extraction while $f_{\phi}$ is a multilayer perceptron (MLP). $f_{\psi}$ is also a MLP for joint decision-making. $\theta,\phi,\psi$ respectively indicates the corresponding parameters of these parts.

As shown in Fig. \ref{FIG2}, it is assumed that the HR images are divided into 4 $\times$ 4 blocks, where each block provides different semantics for the back-end model. The semantic extraction module can choose any combination of them. Therefore, $\textbf{p}$ in Eq. (\ref{eq1}) defines sixteen Bernoulli distributions where $\textbf{p}[i]$ indicates the probability of each block in the image being a semantic at each dimension $i$. The actions performed by the policy network can be sampled at each dimension $\textbf{a}[i]\sim\left\{0,1\right\}^{\textbf{p}[i]}$, where $\textbf{a}[i]=1$ indicates that the $i$-th semantic block is extracted for compression and $\textbf{a}[i]=0$ otherwise. Therefore, the joint probability density function of the output of the policy network can be expressed as
\begin{equation} \label{eq2}
	\pi_{\theta,\phi,\psi}(\textbf{a}|\textbf{X}^{lr},g)=\prod \limits_{i=0}^{15}{{\textbf{p}[i]}^{\textbf{a}[i]}{(1-\textbf{p}[i])}^{1-\textbf{a}[i]}}.
\end{equation}
Then we would model the policy network as a single-step MDP, and the detailed optimization process will be introduced in Sec. \ref{section4}.

\subsection{Block-wise Deep CS}

Compressed sensing, also called sparse sampling, is a technique for finding sparse solutions of underdetermined linear systems, which has been widely used in the field of signal processing in the past two decades. It maps a linear signal to the compressed domain through a sampling matrix. For high-dimensional image signals, the sampling procedure is performed block by block:
\begin{equation} \label{eq3}
	\textbf{y}_i=\boldsymbol{\Phi}{\textbf{x}_i},
\end{equation}
where $\textbf{x}_i\in{{\mathbb R}^n}$ is the vector form of $i$'th sub-block in pixel domain. $\textbf{y}_i\in{{\mathbb R}^m}$ is the compressed signal of $\textbf{x}_i$. $\boldsymbol\Phi\in{{\mathbb R}^{m\times{n}}}(m\ll{n})$ denotes a sampling matrix, which is implemented by a learnable convolution kernel in our system. Note that the sub-block in CS is different from the semantic blocks introduced in previous section.

Traditional methods of recovering the signal $\textbf{x}_i$ need to solve a convex optimization problem of $l_1$-norm, which is usually of high complexity \cite{33}. In recent years, with the rise of deep learning, some researchers have used CNN to reconstruct the signal $\textbf{x}_i$ \cite{34}. Since the sampling matrix can also be implemented through a convolutional layer, both compression and reconstruction steps can be trained through an end-to-end CNN. In our system, $\textbf{x}_i\in{\mathbb R}^{k\cdot{k}\cdot3}$ is a vector reshaped from an image block $\textbf{X}_i\in{\mathbb R}^{k\times{k}\times3}$ with channels RGB, and $k$ denotes the height and width of the sampled sub-blocks. We use a kernel to convolve the image:
\begin{equation} \label{eq4}
	\textbf{Y}=\boldsymbol\kappa\star\textbf{X},
\end{equation}
where $\boldsymbol\kappa\in{\mathbb R}^{k\times{k}\times3\times{m}}$ is the convolution kernel without bias, $m=\lceil{k}\cdot{k}\cdot3\cdot{sr}\rceil$, $sr$ is the sampling rate. $\textbf{X}=stitch\{\textbf{X}_1,\textbf{X}_2,\cdots,\textbf{X}_P\},\textbf{Y}=stitch\{\textbf{Y}_1,\textbf{Y}_2,\cdots,\textbf{Y}_P\}$. $\star$ indicates a convolution operation with a stride of k. $stitch$ means to stitch all image blocks into a whole image. $P$ is the total number of image blocks. The elements in the convolution kernel $\boldsymbol{\kappa}$ and the sampling matrix $\boldsymbol{\Phi}$ are one-to-one correspondence. The two procedures described by Eq. (\ref{eq3}) and Eq. (\ref{eq4}) are equivalent.

A denoising-based approximate message passing algorithm (D-AMP) was proposed in \cite{33}, which alternately iterates the residual variables and the reconstructed signal through a denoising function:
%\begin{equation} \label{eq5}
%	\begin{cases}
%		\textbf{z}^{l}=\textbf{y}-\mathbf{\boldsymbol{\Phi}}\textbf{x}^{l}+\textbf{z}^{l-1}{\rm{div}}D_{\hat{\sigma}^{l-1}}{(\textbf{x}^{l-1}+{\boldsymbol{\Phi}}^{-}\textbf{z}^{l-1})}/m\\
%		\textbf{x}^{l+1}=D_{\hat{\sigma}^{l}}\left(\textbf{x}^{l}+\mathbf{\Phi}^{-}\textbf{z}^{l}\right)
%	\end{cases}
%\end{equation}
\begin{equation} \label{eq5}
	\textbf{z}^{l}=\textbf{y}-\mathbf{\boldsymbol{\Phi}}\textbf{x}^{l}+\textbf{z}^{l-1}{\rm{div}}D_{\hat{\sigma}^{l-1}}{(\textbf{x}^{l-1}+{\boldsymbol{\Phi}}^{-1}_{\rm{left}}\textbf{z}^{l-1})}/m,
\end{equation}
\begin{equation} \label{eq6}
	\textbf{x}^{l+1}=D_{\hat{\sigma}^{l}}\left(\textbf{x}^{l}+{\boldsymbol{\Phi}}^{-1}_{\rm{left}}\textbf{z}^{l}\right),
\end{equation}
where $\textbf{z}^{l}$ is the iterative residual in the compressed domain, $D_{\hat{\sigma}}$ is a Gaussian denoising filter with parameter $\hat{\sigma}$. ${\rm{div}}D_{\hat{\sigma}^{l-1}}{(\cdot)}$ is the divergence of the output Gaussian distribution for the correction of the residual. ${\boldsymbol{\Phi}}^{-1}_{\rm{left}}$ is the left inverse matrix of $\boldsymbol\Phi$. 

\begin{figure*}[!h]
	\centering
	\includegraphics{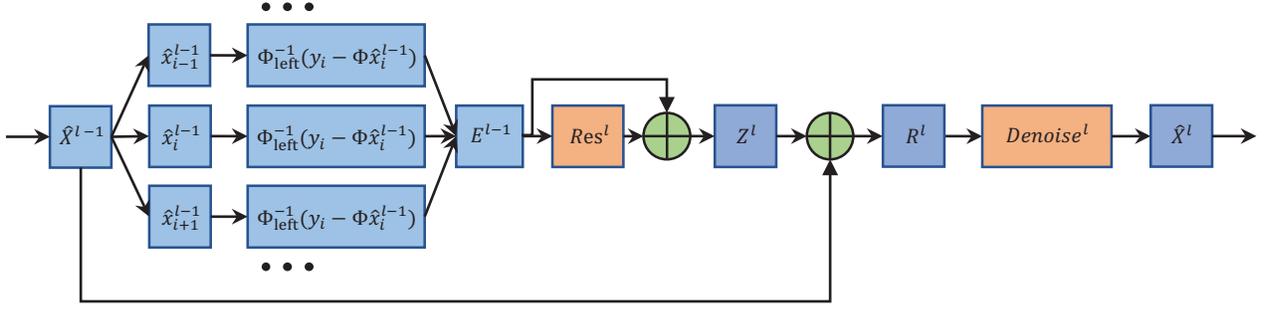}
	\caption {Diagram of block-wise deep reconstruction at $t$'th stage.}
	\vspace{0cm}
	\label{FIG3}
\end{figure*}

In order to use CNN to replace the residual correction term to avoid complicated divergence calculations, we can use a residual structure CNN to predict a new residual from the original term, as illustrated in Fig. \ref{FIG3}. Suppose that the residual consists of $L$ iterations, in each iteration $\hat{\textbf{X}}^{l-1}$ represents the reconstructed image output from the previous stage. The residuals for each sub-block in the compressed domain $\{\textbf{y}_1-\boldsymbol\Phi\hat{\textbf{x}}_1^{l-1},\textbf{y}_2-\boldsymbol\Phi\hat{\textbf{x}}_2^{l-1},\cdots,\textbf{y}_N-\boldsymbol\Phi\hat{\textbf{x}}_N^{l-1}\}$ can be calculated through the sampling matrix. The residual iteration in the compressed domain may still cause serious blocking effects. Mapping the residuals back to the pixel domain with ${\boldsymbol{\Phi}}^{-1}_{\rm{left}}$ for correction can solve this problem:
%\begin{equation}\label{eq7}
%	\begin{aligned}
%		\begin{split}
%			E^{l-1}&=stitch\{\Phi^-(y_1-\Phi\hat{x}_1^{l-1}),\Phi^L(y_2-\Phi\hat{x}_2^{l-1}),\\
%					&\cdots,\Phi^-(y_N-\Phi\hat{x}_N^{l-1})\}\\
%			&=\kappa^\star\overline{\ast}{(Y-\kappa\ast{\hat{X}^{l-1}})},
%		\end{split}
%	\end{aligned}
%\end{equation}
\begin{equation}\label{eq7}
	\textbf{e}_i^{l-1}={\boldsymbol{\Phi}}^{-1}_{\rm{left}}(\textbf{y}_i-\boldsymbol\Phi\hat{\textbf{x}}_i^{l-1}),
\end{equation}
\begin{equation}\label{eq8}
	\textbf{E}^{l-1}=\boldsymbol\kappa^{-1}\ \overline{\star}\ {(\textbf{Y}-\boldsymbol\kappa\star{\hat{\textbf{X}}^{l-1}})},
\end{equation}
where $\textbf{e}_i^{l-1}$ denotes the residual vector of $i$'th image block at $(l-1)$'th stage,  $\textbf{E}^{l-1}=stitch\{\textbf{E}_1^{l-1},\textbf{E}_2^{l-1},\cdots,\textbf{E}_P^{l-1}\}$ is the residual image at $(l-1)$'th stage. $\boldsymbol\kappa^{-1}\in{\mathbb R}^{k\times{k}\times{m}\times3}$ represents the kernel corresponding to $\boldsymbol\Phi^{-1}_{\rm{left}}$, $\overline{\star}$ indicates the transposed convolution with a stride of $k$.

Then update the residual in the pixel domain
\begin{equation}\label{eq9}
	\textbf{Z}^l=\textbf{E}^{l-1}+f_{\omega^l}(\textbf{Z}^{l-1}),
\end{equation}
where $\boldsymbol\omega^l$ is the parameters of the residual network with two convolutional layers. $\textbf{Z}^l$ represents the updated residuals at $l$-th stage. 

Finally, the image $\hat{\textbf{X}}^l$ recovered at this stage is output by the denoising network:
\begin{equation}\label{eq10}
	\hat{\textbf{X}}^l=f_{\boldsymbol\Omega^l}{(\hat{\textbf{X}}^{l-1}+\textbf{Z}^l)},
\end{equation}
where $\boldsymbol\Omega^l$ indicates the parameters of the denoising network with three convolutional layers.

Since only the semantic part of the image is transmitted, but the reconstruction procedure is performed on the entire image, the compressed values of the non-semantic part is set to 0 to reduce the computational complexity of the reconstruction network. Therefore, the parameters of all layers do not contain bias.

\subsection{Image Transmission Model}
The transmission latency of the system is mainly determined by the transmission rate and the number of bits transmitted. The transmission latency of the uplink is considered in the proposed system. We assume that the channel between the drones and the BS follows the stationary Rayleigh fading distribution, where the channel gain and transmission rate remain static within a small time interval. Then, the uplink transmission rate at time slot $t$ is as follows:
\begin{equation}\label{eq11}
	r_{D\rightarrow{B}}(t)=w_{D\rightarrow{B}}\log_2(1+\frac{p(t) g(t)}{N_0}),
\end{equation}
where $w_{D\rightarrow{B}}$ denotes the BS bandwidth allocated to drones. The signal-to-interference-plus-noise ratio (SINR) for the signals received at the BS is expressed as $\rm{SINR}={p(t) g(t)}/{N_0}$, where $p(t)$ is the transmission power of drones. The additive white Gaussian noise (AWGN) is $N_0$. The channel gain for the signals received at the BS is $g(t)=d^{-\alpha}u_t$ for instantaneous perception of communication environment. The distance $d$ between drones and BS is the most important factor affecting the channel gain. We will consider the path loss exponent $\alpha$ and Rayleigh distribution $u_t$ with unit mean. Generally, channel gain is an unlimited continuous variable. It can be quantified in $g(t)\in$\{-30 dB, -20 dB, -10 dB, 0 dB, 10 dB, 20 dB, 30 dB\}, making it easier for the model to discriminate.

Assuming that the policy network has extracted a total of $N^a$ semantic blocks, the number of bits required to be transmitted is given by
\begin{equation}\label{eq12}
	N^b=\gamma\cdot(\frac{H}{4})\cdot(\frac{W}{4})\cdot{N^a}\cdot{sr},
\end{equation}
where $H$ and $W$ represent the height and width of the HR images. $\gamma$ implies number of quantization bits in compressed domain.

Then the uplink transmission latency is given by
\begin{equation}\label{eq13}
	T_{up}=\mathop{\arg\min}\limits_{\int_0^\delta{r}(t)dt\geq{N^b}}{\delta}\approx\frac{N^b}{r_{D\rightarrow{B}}(t)},
\end{equation}

As mentioned above that the transmission rate $r_{D\rightarrow{B}}(t)$ in each time slot under a stationary channel is constant, the latency is approximately the ratio of the number of transmitted bits $N^b$ to $r_{D\rightarrow{B}}(t)$.

\section{System Optimization}
\label{section4}
In this section, we will give a detailed introduction of the optimization procedure of the policy network and the deep CS network for semantic extraction respectively.

\subsection{Optimization of Policy Network}
As described in Sec. \ref{section3}, the policy network is modeled as a single-step MDP. At any time, the image captured by the sensor and the channel gain sensed in real time are two independent variables. For the proposed model, the image is a high-dimensional continuous variable while the channel gain is a single discrete value. The output policy $\textbf{p}$ implies $2^{16}$ discrete actions that may be performed by the agent. It is tricky to optimize our high-dimensional system with traditional reinforcement learning methods based on value functions. Policy gradient is not only suitable for our high-dimensional system, but also capable of leanring random policies that cannot be learned by the value function. The purpose is to optimize the parameters $\theta,\phi,\psi$ by maximizing the expected value of cumulative rewards on all policy trajectories:
\begin{equation}\label{eq14}
	\pi_{\theta,\phi,\psi}^{*}=\underset{\theta,\phi,\psi}{\operatorname{argmax}} \mathbb{E}_{\tau \sim \pi_{\theta,\phi,\psi}(\tau)}[r(\tau)],
\end{equation}
where $\tau\sim\pi(\tau)$ is a trajectory of states, actions, and rewards sampled from an interaction with our policy, and $r(\tau)$ represents the sum of all rewards for this trajectory.

For our single-step MDP process, trajectory $\tau$ is composed of $\textbf{X}^{lr},g$ and $\textbf{a}$. The objective is to maximize the accuracy of the back-end target model while minimizing the uplink transmission latency of the system by maximizing a reward function:
\begin{equation}\label{eq15}
	J\left(\theta,\phi,\psi\right)=\frac{1}{MN} \sum_{m=1}^{M} \sum_{n=1}^{N} \log \pi_{\theta,\phi,\psi} \left(\textbf{a}_{m,n} \mid \textbf{X}^{lr}_m,g_n\right) R_{m,n}
\end{equation}
where $M$ and $N$ respectively represent the total number of $\textbf{X}^{lr}_m$ and $g_n$. $R_{m,n}$ indicates the single-step reward from the system with the input of $\textbf{X}^{lr}_m$ and $g_n$, which is defined by the system's transmission latency $T_{m,n}$ and the correctness of the target model
\begin{equation}\label{eq16}
	R_{m,n}=
	\begin{cases}
		\mathcal{R}(T_{m,n}), & \mathop{\arg\max}\limits_{k}f_k^{tar}(\hat{\textbf{X}}_{m}^L)=C \\
		\eta, & else
	\end{cases},
\end{equation}
where $f_k^{tar}$ denotes the $k$-th output of the backend target model. $C$ is the label of input $\textbf{X}$, and $\mathcal{R}(T_{m,n})$ is a function negatively correlated to $T_{m,n}$. A penalty term is represented by $\eta$. Two kinds of $\mathcal{R}(T_{m,n})$ are tried in proposed system
\begin{equation}\label{eq17}
	\mathcal{R}(T_{m,n})=
	\begin{cases}
		\frac{1}{1+\lambda{T_{m,n}}} \\
		\exp{(-\lambda{T_{m,n}}})
	\end{cases},
\end{equation}
where $\lambda$ is a positive hyperparameter.

The gradients of $\theta,\phi,\psi$ can be updated along the gradient ascent. To avoid over-update of the network with great rewards and keep training stable, an offset need to be subtracted:
\begin{equation}\label{eq18}
	\nabla_{\theta,\phi,\psi}J=\frac{1}{MN} \sum_{i=1}^{M} \sum_{j=1}^{N} \nabla_{\theta,\phi,\psi} \pi\left(\textbf{a}_{m,n} \mid \textbf{X}^{lr}_m,g_n\right) (R_{m,n}-\overline{R}_{m,n}),
\end{equation}
where $\overline{R}_{m,n}$ is the system reward when performing a baseline action $\overline{\textbf{a}}$, which directly selects the image blocks according to the larger value of the output policy without sampling
%\begin{equation}\label{eq19}
%	\overline{\textbf{a}}_{m,n}[i]=\mathop{\arg\max}\limits_{\textbf{a}}\pi\left(\textbf{a}_{m,n}[i] \mid \textbf{X}^{lr},g_j\right).
%\end{equation}
\begin{equation}\label{eq19}
	\overline{\textbf{a}}_{m,n}[i]=
	\begin{cases}
		1, & \textbf{p}[i] \geq 1-\textbf{p}[i] \\
		0, & else
	\end{cases}.
\end{equation}

Since the parameters of the policy network are divided into three parts, and the two input variables are independent of each other, we use the control variable method to train it. This optimization process is divided into three stages. First, the channel gain is kept changing and $\phi$ are kept constant in the outer loop, while $\theta$ and $\psi$ are updated with different LR images in the inner loop. In the middle stage, the LR images are kept changing and $\theta$ are kept constant in the outer loop, while $\phi$ and $\psi$ are updated with different channel gains in the inner loop. In the last stage, $\theta,\phi,\psi$ are updated synchronously with the LR images and channel gains being sampled randomly.

\subsection{Optimization of Deep CS Model}

As illustrated in Sec. \ref{section3}, the sampling and compression processes of CS are carried out simultaneously. By deploying the sampling matrix $\boldsymbol{\Phi}$ in the front-end UAV, the measured values of semantic blocks in compressed domain can be obtained quickly. The measurement matrices of traditional CS algorithms are usually Gaussian matrices or Hadamard matrices, which should be constructed to satisfy Restricted Isometry Property (RIP) conditions. In proposed system, $\boldsymbol{\Phi}$ can be implemented through a convolutional layer and an optimal parameter within the kernel can be learned over the data distribution. Because the deep reconstruction procedure requires the auxiliary calculation of the inverse matrix $\boldsymbol{\Phi}^{-1}_{\rm{left}}$, it is necessary to pretrain the sampling matrix $\boldsymbol{\Phi}$ through an auxiliary layer due to the underivativity of the inversion operation. Assuming that the transposed convolution kernel of the auxiliary layer is $\kappa^\prime\in{\mathbb R}^{k\times{k}\times{m}\times3}$ with a stride of $k$, an initial reconstruction process can be given by 
\begin{equation} \label{eq20}
	\hat{\textbf{X}}^0=\boldsymbol\kappa^\prime\overline{\star}{\textbf{Y}},
\end{equation}
where Eq. (\ref{eq4}) and Eq. (\ref{eq20}) compose a pair of compressing and preliminary reconstruction process. The pre-reconstructed image can be approximated by minimizing the mean square error (MSE) loss between $\textbf{X}$ and $\hat{\textbf{X}}^0$:
\begin{equation} \label{eq21}
	\hat{\kappa}=\mathop{\arg\min}\limits_{\kappa}\frac{1}{2M}\sum_{m=1}^{M} ({\boldsymbol\kappa}^\prime\overline{\star} (\boldsymbol\kappa\star{\textbf{X}_m})-\textbf{X}_m)^2
\end{equation}

Once the optimal sampling kernel $\boldsymbol\kappa$ is learned, $\boldsymbol\kappa^{-1}$ can be calculated according to the inverse matrix $\boldsymbol{\Phi}_{\rm{left}}^{-1}$. The pre-reconstructed image $\hat{\textbf{X}}^0$ for each HR image can be calculated through this auxiliary, which can be used to optimize the deep reconstruction layers in the subsequent stages. Given the set of original images and reconstructed images $\{\textbf{X}_m, \hat{\textbf{X}}_m^L\}_{m=1}^{M}$, the parameters in multi-stage reconstruction layers can also be updated with MSE loss:
\begin{equation} \label{eq22}
	\hat{\omega}^l,\hat{\Omega}^l=\mathop{\arg\min}\limits_{\omega^l,\Omega^l}\frac{1}{2M}\sum_{m=1}^{M} (\hat{\textbf{X}}_m^L-\textbf{X}_m)^2.
\end{equation}

Although only semantic blocks are compressed and transmitted, in order to avoid the block effect caused by reconstruction, the training process of the entire reconstruction network is carried out on the full images, without any special operations for any macroblock. Since the reconstructed network does not contain bias, setting the measurement values of the non-semantic regions to 0 during reconstruction will not affect the reconstruction quality too much.

When optimizing the entire system, the two modules are updated in an alternate iterative manner. First, the deep CS model is trained with the full HR images. Afterwards, the semantic extraction model can be trained by this model, since the calculation of system rewards depends on the quality of compression reconstruction. Driven by policy network trained at the previous stage, the deep CS model can be finetuned with HR images.

\section{Simulation Results and Discussion}
In this section, we evaluate the performance of the proposed task oriented algorithm for joint image compression and classification with computer simulations. In addition, we compare the proposed algorithm with other six manual designed and traditional content perception policies.

\subsection{Simulation Setup}
In this section, the experimental setup for the proposed scheme has been introduced. The experiments are conducted in an Ubuntu operating system (CPU Intel core i7-10700 2.9 GHz; memory 64GB, GPU NVIDIA GeForce RTX 3090, which contains 10496 CUDA computing core units and 24GB graphics memory). We use Python and deep learning framework of Pytorch for simulation.

The transmission power of the drones, the noise of the channel and the bandwidth are fixed to 15 dBm, -97 dB, and 100 KHz, respectively. We divide the channel gains into 7 states as the input of the perception network: $g(t)\in\{-30,-20,-10,1,10,20,30\}$ dB. As a result, their corresponding 7 uplink transmission rates are $r_{D\rightarrow{B}}(t)\in\{105,355,675,1006,1338,1670,2003\}$ kbps.

We evaluated the performance of the proposed algorithm on the Aerial Image Dataset (AID) dataset \cite{35} which has 10000 images with 30 different scene classes of size $600 pt \times 600 pt$. The 30 aerial scene types are as follows: airport, bare land, baseball field, beach, bridge, center, church, commercial, dense residential, desert, farmland, forest, industrial, meadow, medium residential, mountain, park, parking, playground, pond, port, railway station, resort, river, school, sparse residential, square, stadium, storage tanks and viaduct. Each class contains 220 to 420 samples. $N_{train}=8000$ samples are randomly selected as the training set with the other $N_{test}=2000$ samples as the test set. These two sets are consistent during training and testing stages of all aforementioned modules. The target modeled deployed on the back-end server is a ResNet34 with the input size of $224\times224$. Therefore, all HR images are resized to this size. The size of each semantic block is $56\times56$. The receptive field of proposed deep CS model compress is $k=16$. The sampling rate $sr$ is set to 0.3 for all of the semantic blocks. All experiments use 8-bit quantization in the compressed domain by default. 

\subsection{Simulation Results}

In the experimental environment set up in the previous section, the performance of two reward functions have been tested, as shown in Tab. \ref{TAB1}. The first two rows of Tab. \ref{TAB1} indicate 7 channel states and their corresponding transmission rates respectively. The third row means the transmission latency of full HR images in the compressed domain. The two reward functions are $\mathcal{R}_1=1/(1+2{T}),\eta_1=0.15$ and $\mathcal{R}_2=\exp(-2T),\eta_2=-0.02$. $\overline{N}^a=1/N_{test}\sum_{i=1}^{N_{test}} N_i^a$ represents the average number of semantic blocks extracted and transmitted by the two policies learned under different channel conditions. $\overline{B}=1/N_{test}\sum_{i=1}^{N_{test}} B_i$ indicates the average semantic data transmitted by the two policies under different channel conditions. $\overline{T}=\overline{B}/r_{D\rightarrow{B}}$ is the average transmission latency in these 7 states.

It can be seen from Tab. \ref{TAB1} that the proposed method can extract different amounts of blocks for transmission according to channel conditions. Additionally, the proposed policy learning method has been compared with 6 other policies under  the premise of transmitting the same number of semantic blocks. Assuming that under the gain index $j\in\{1,2,3,4,5,6,7\}$, there are $N_j$ samples in the $N_{test}$ transmitted samples can be correctly classified by the back-end model by the proposed method, then its accuracy is calculated by $N_j/N_{test}$. Assuming that under the channel condition $j$, the number of semantic blocks transmitted by the $i$-th test sample is $N_{i,j}^a$, then other policies are controlled to also transmit $N_{i,j}^a$ semantic blocks, which can ensure the fairness of the experiment.

\begin{table*}[!h]
	\centering
	\caption{Experimental reults with two reward functions.}
	\begin{tabular}{c|cp{4.19em}ccccccc}
		\toprule
		\multicolumn{1}{c}{} & \multicolumn{1}{c}{$s^c$(dB)} & \multicolumn{1}{c}{} & -30   & -20   & -10   & 0   & 10   & 20   & 30 \\
		\multicolumn{1}{c}{} & \multicolumn{1}{c}{$r_{D\rightarrow{B}}$(kbps)} & \multicolumn{1}{c}{} & 105   & 355   & 675   & 1006  & 1338  & 1670  & 2003 \\
		\multicolumn{1}{c}{} & \multicolumn{1}{c}{$T_{all}$(ms)} & \multicolumn{1}{c}{} & 3438  & 1018  & 535   & 359   & 270   & 216   & 180 \\
		\midrule
		\multicolumn{1}{c|}{\multirow{10}[3]{*}{$\mathcal{R}_1,\mathcal{P}_1$}} & \multicolumn{2}{c}{$\overline{N}^a$} & 9.129 & 9.273 & 10.05 & 10.727 & 11.311 & 11.36 & 11.735 \\
		& \multicolumn{2}{c}{$\overline{B}$(kB)} & 25.766 & 26.172 & 28.365 & 30.276 & 31.924 & 32.062 & 33.121 \\
		& \multicolumn{2}{c}{$\overline{T}$(ms)} & 1962  & 590   & 336   & 241   & 191   & 154   & 132 \\
		\cmidrule{2-10}          & \multicolumn{1}{c}{\multirow{7}[1]{*}{Accuracy}} & proposed & 0.822 & 0.831 & 0.867 & 0.893 & 0.912 & 0.912 & 0.919 \\
		&       & policy1 & 0.709 & 0.72  & 0.757 & 0.794 & 0.825 & 0.826 & 0.847 \\
		&       & policy2 & 0.604 & 0.617 & 0.689 & 0.736 & 0.772 & 0.775 & 0.807 \\
		&       & policy3 & 0.786 & 0.801 & 0.839 & 0.863 & 0.884 & 0.886 & 0.9 \\
		&       & policy4 & 0.781 & 0.793 & 0.834 & 0.866 & 0.883 & 0.885 & 0.897 \\
		&       & policy5 & 0.314 & 0.307 & 0.419 & 0.488 & 0.529 & 0.546 & 0.582 \\
		&       & policy6 & 0.365 & 0.38  & 0.461 & 0.54  & 0.589 & 0.588 & 0.62 \\
		\midrule
		\multicolumn{1}{c|}{\multirow{10}[3]{*}{$\mathcal{R}_2,\mathcal{P}_2$}} & \multicolumn{2}{c}{$\overline{N}^a$} & 9.594 & 9.703 & 10.27 & 11.029 & 11.442 & 11.57 & 12.522 \\
		& \multicolumn{2}{c}{$\overline{B}$(kB)} & 27.077 & 27.384 & 28.986 & 31.128 & 32.294 & 32.654 & 35.342 \\
		& \multicolumn{2}{c}{$\overline{T}$(ms)} & 2061.38 & 617.019 & 343.174 & 247.368 & 192.976 & 156.336 & 141.146 \\
		\cmidrule{2-10}          & \multicolumn{1}{c}{\multirow{7}[2]{*}{Accuracy}} & proposed & 0.833 & 0.839 & 0.87  & 0.901 & 0.911 & 0.914 & 0.931 \\
		&       & policy1 & 0.717 & 0.72  & 0.765 & 0.811 & 0.837 & 0.844 & 0.884 \\
		&       & policy2 & 0.632 & 0.64  & 0.689 & 0.752 & 0.789 & 0.796 & 0.846 \\
		&       & policy3 & 0.802 & 0.805 & 0.846 & 0.875 & 0.892 & 0.894 & 0.92 \\
		&       & policy4 & 0.793 & 0.799 & 0.833 & 0.869 & 0.885 & 0.889 & 0.913 \\
		&       & policy5 & 0.376 & 0.389 & 0.457 & 0.535 & 0.58  & 0.597 & 0.691 \\
		&       & policy6 & 0.423 & 0.437 & 0.496 & 0.575 & 0.609 & 0.621 & 0.71 \\
		\bottomrule
	\end{tabular}%
	\label{TAB1}%
\end{table*}%

\begin{figure}[!h]
	\centering
	\subfigure[$\mathcal{R}_1=1/(1+2{T}),\eta_1=0.15$]{
		\label{fig4:subfig:a}
		\includegraphics[width=7.4cm]{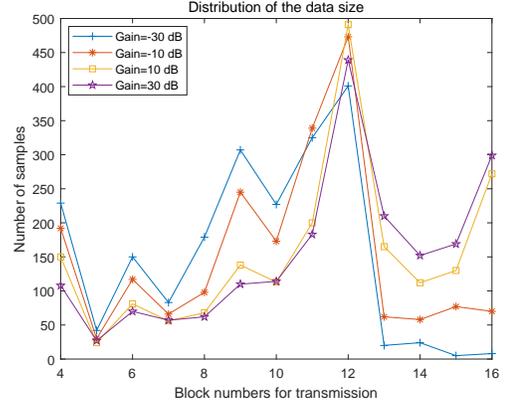}}
	\subfigure[$\mathcal{R}_2=\exp(-2T),\eta_2=-0.02$]{
		\label{fig4:subfig:b}
		\includegraphics[width=7.4cm]{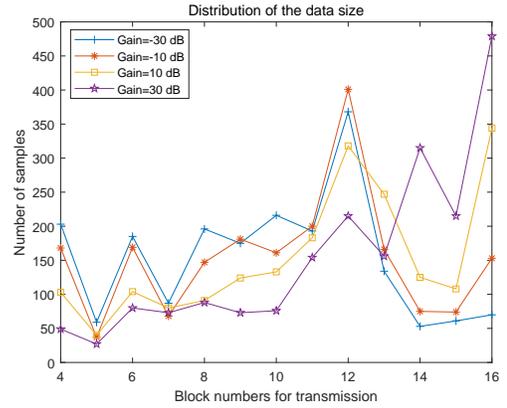}}
	\caption {The number distribution of transmitted blocks under different channel conditions with two different reward functions compared to other policies.}
	\label{FIG4}
\end{figure}

\begin{figure*}[!h]
	\centering
	\subfigure[6 transimission blocks]{
		\label{fig5:subfig:a} %% label for first subfigure
		\includegraphics[width=4.2cm]{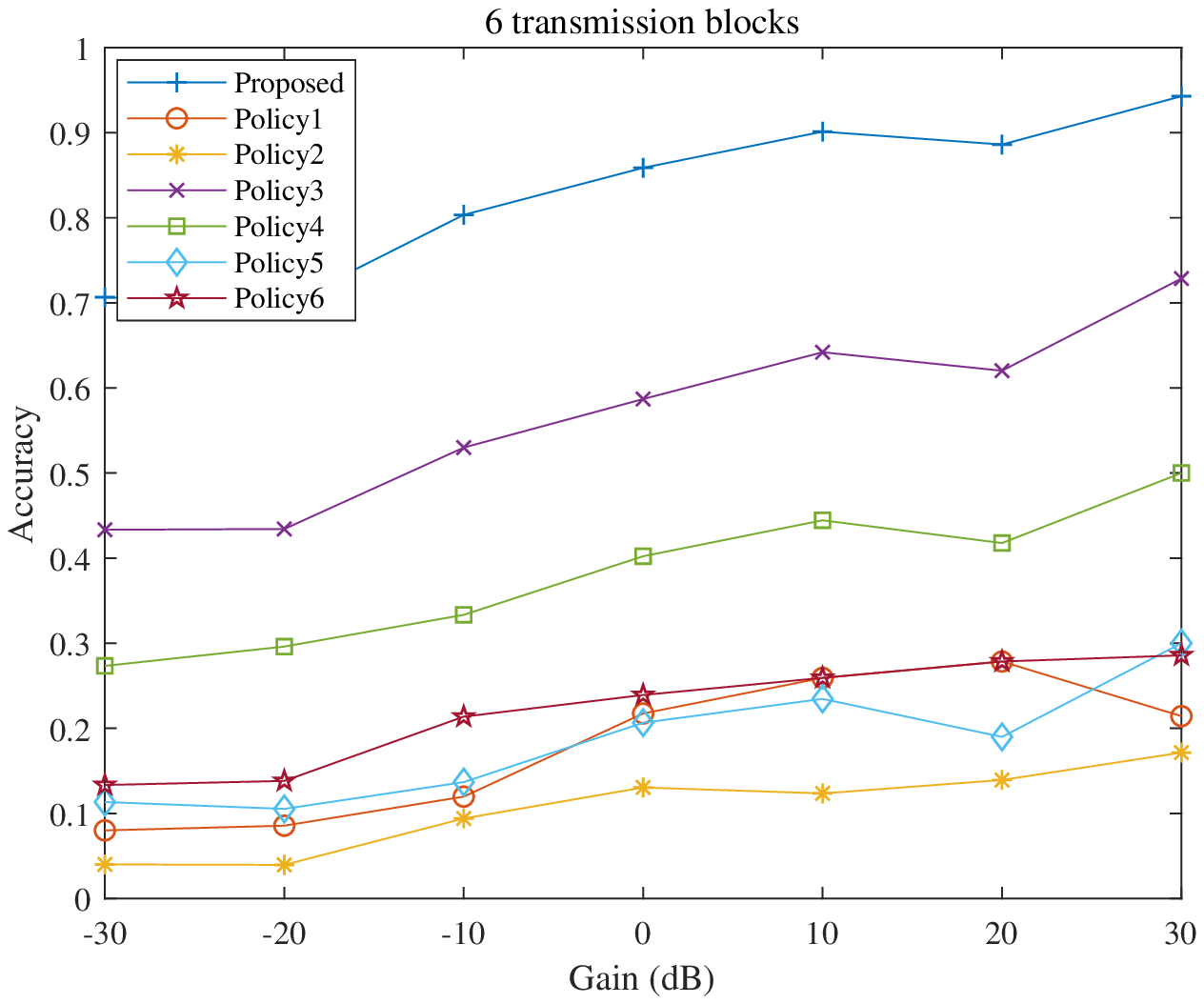}}
	\subfigure[8 transimission blocks]{
		\label{fig5:subfig:b} %% label for second subfigure
		\includegraphics[width=4.2cm]{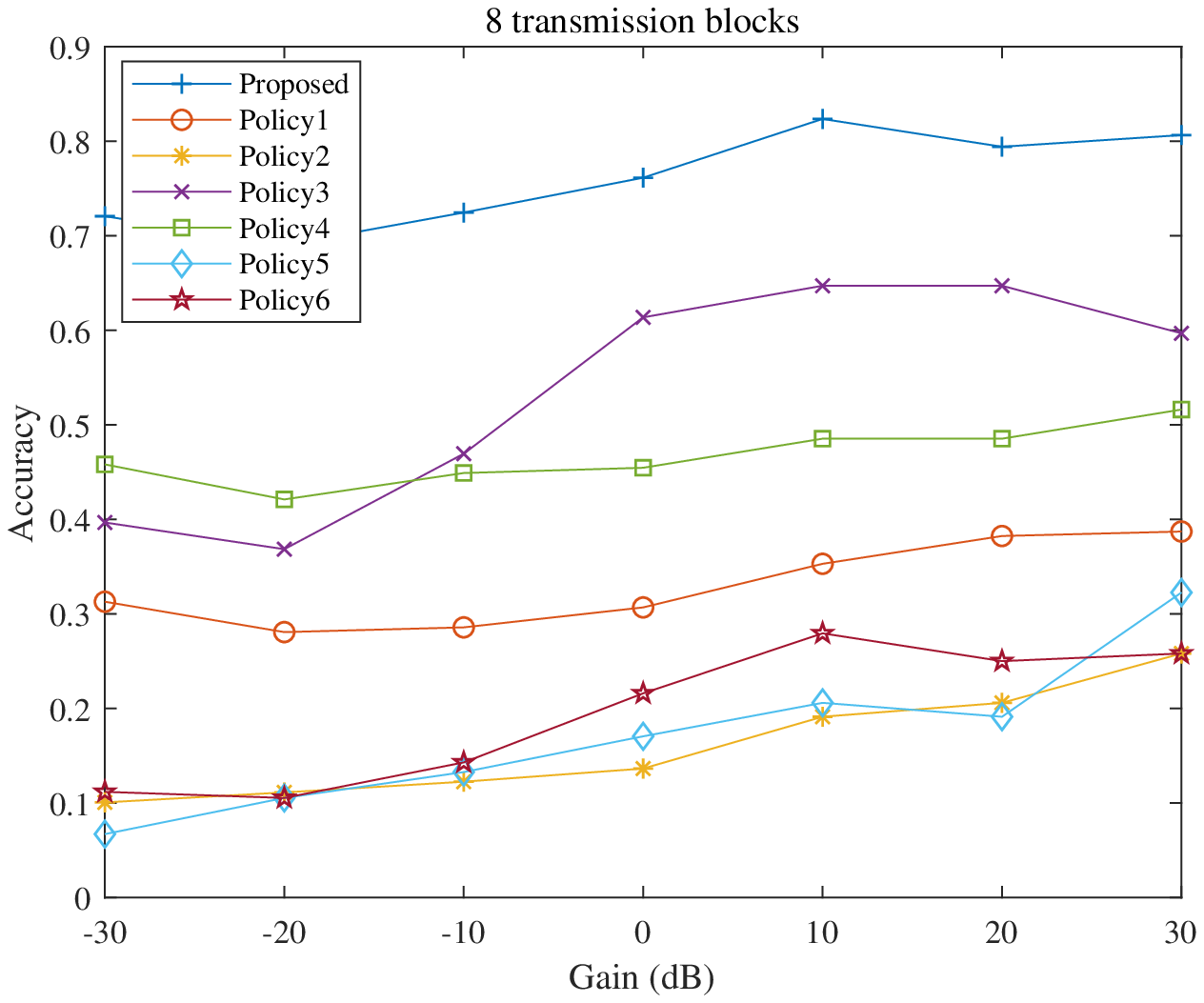}}
	\subfigure[10 transimission blocks]{
		\label{fig5:subfig:c} %% label for second subfigure
		\includegraphics[width=4.2cm]{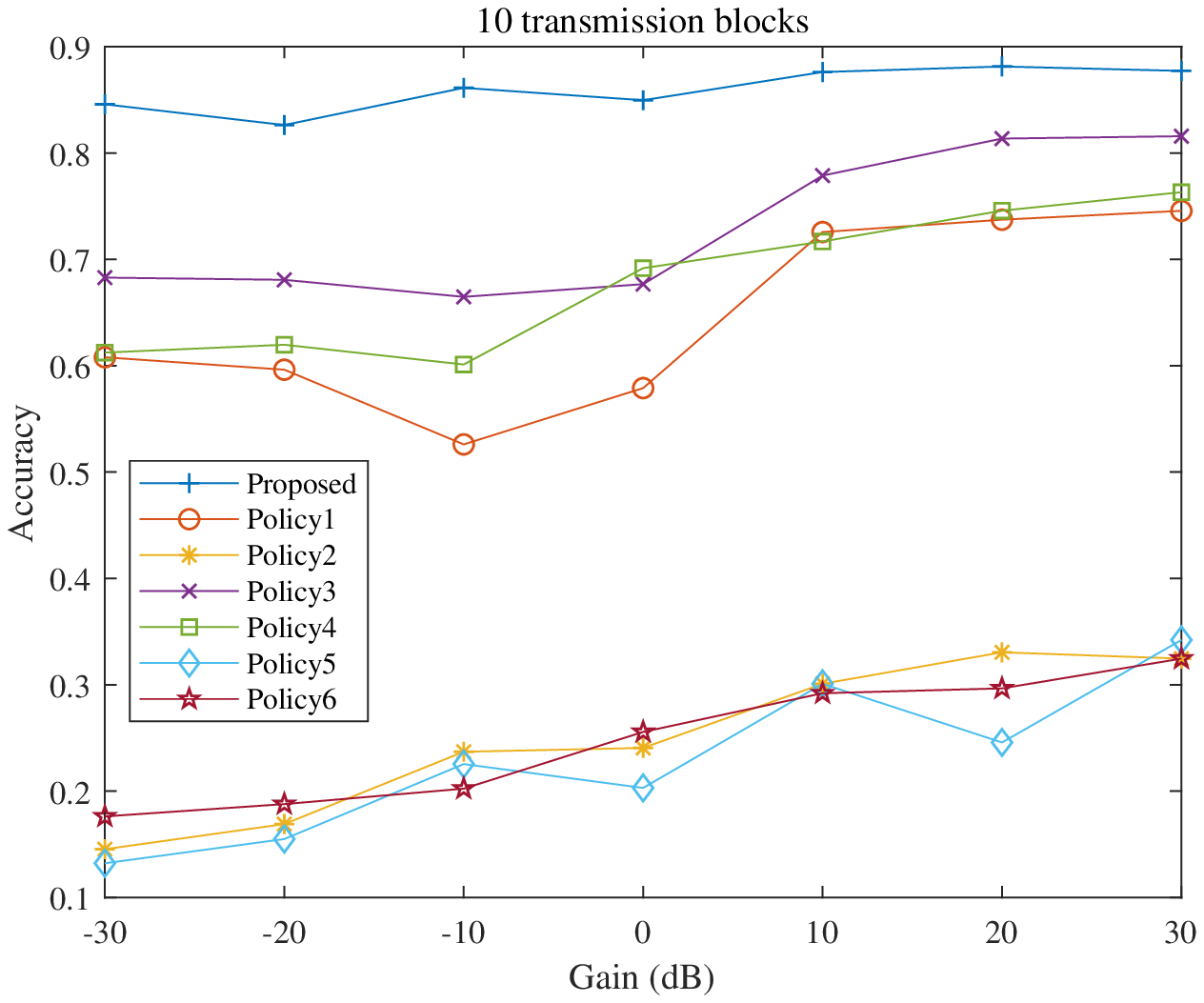}}
	\subfigure[12 transimission blocks]{
		\label{fig5:subfig:d} %% label for second subfigure
		\includegraphics[width=4.2cm]{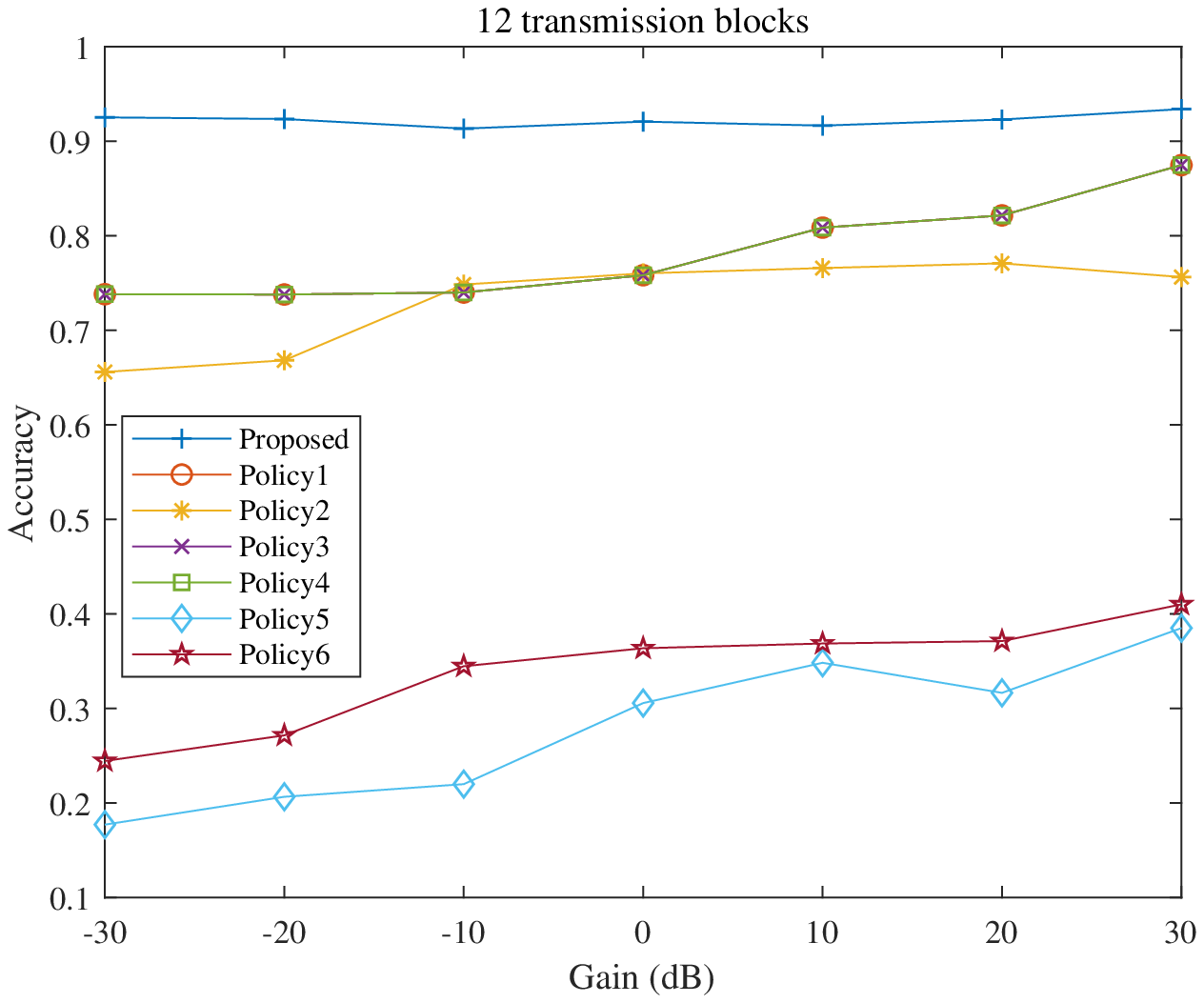}}
	\caption {Back-end accuracy of target model with reward function 1 compared to other policies.}
	\label{FIG5}
\end{figure*}

\begin{figure*}[!h]
	\centering
	\subfigure[6 transimission blocks]{
		\label{fig6:subfig:a} %% label for first subfigure
		\includegraphics[width=4.2cm]{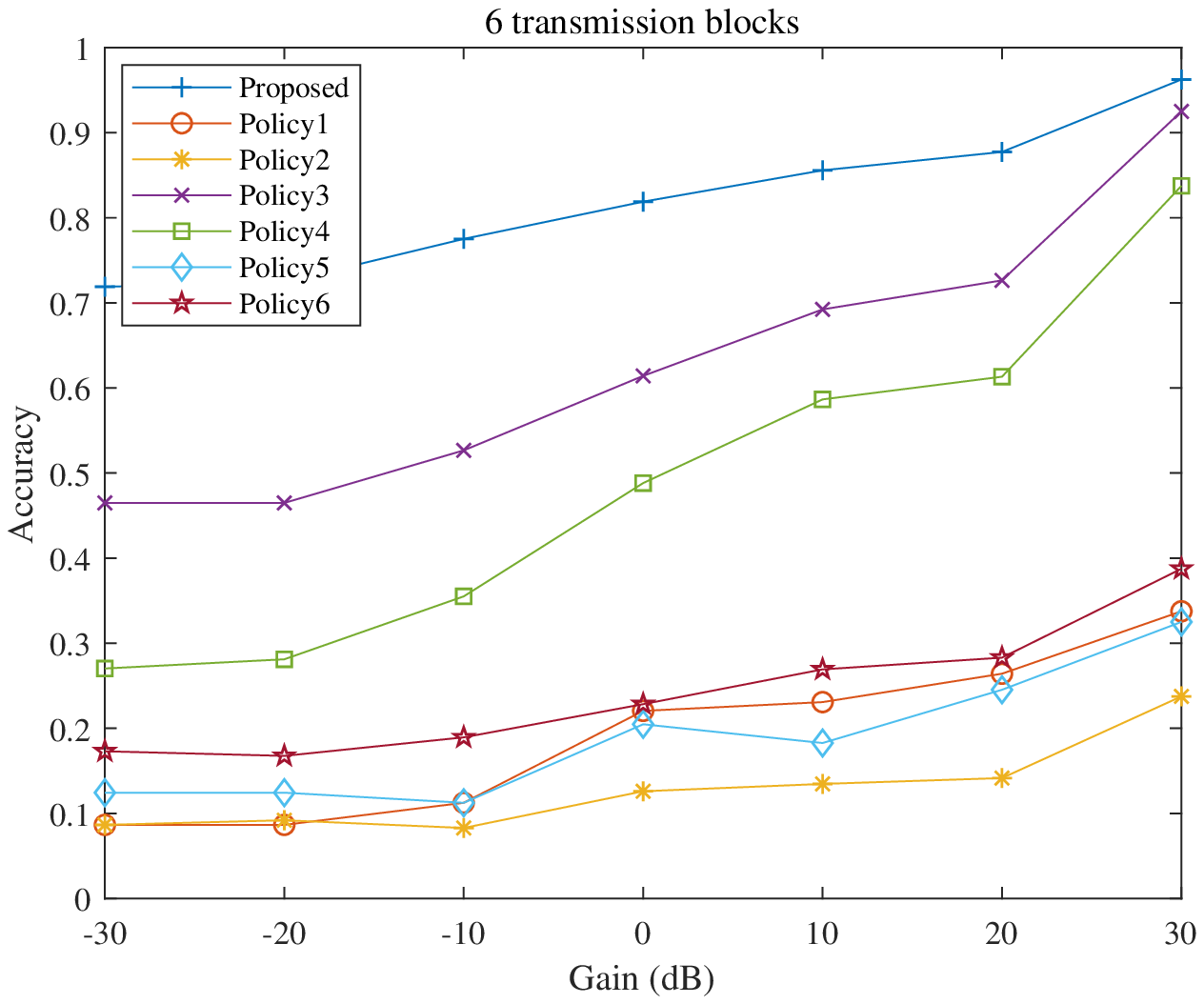}}
	\subfigure[8 transimission blocks]{
		\label{fig6:subfig:b} %% label for second subfigure
		\includegraphics[width=4.2cm]{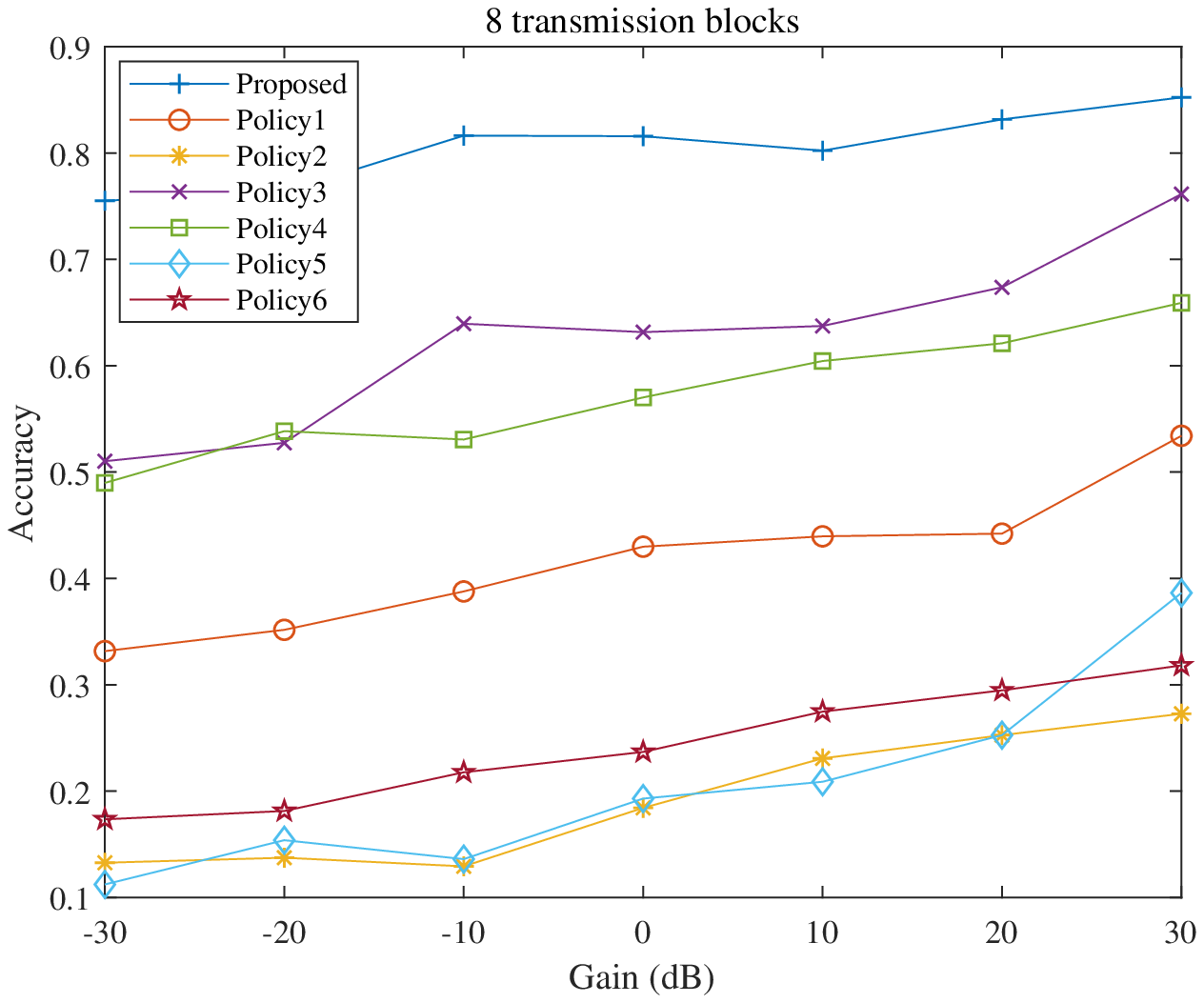}}
	\subfigure[10 transimission blocks]{
		\label{fig6:subfig:c} %% label for second subfigure
		\includegraphics[width=4.2cm]{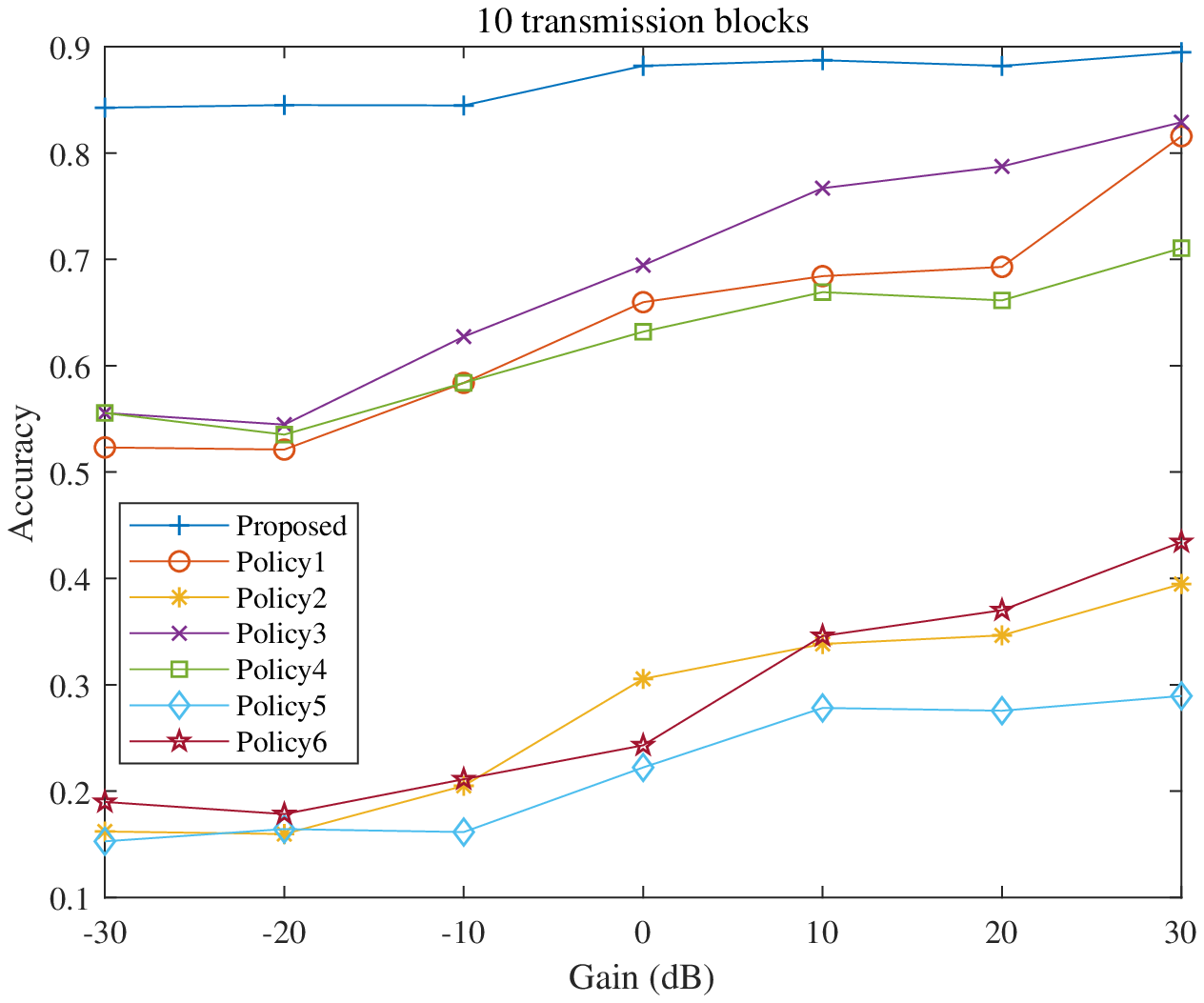}}
	\subfigure[12 transimission blocks]{
		\label{fig6:subfig:d} %% label for second subfigure
		\includegraphics[width=4.2cm]{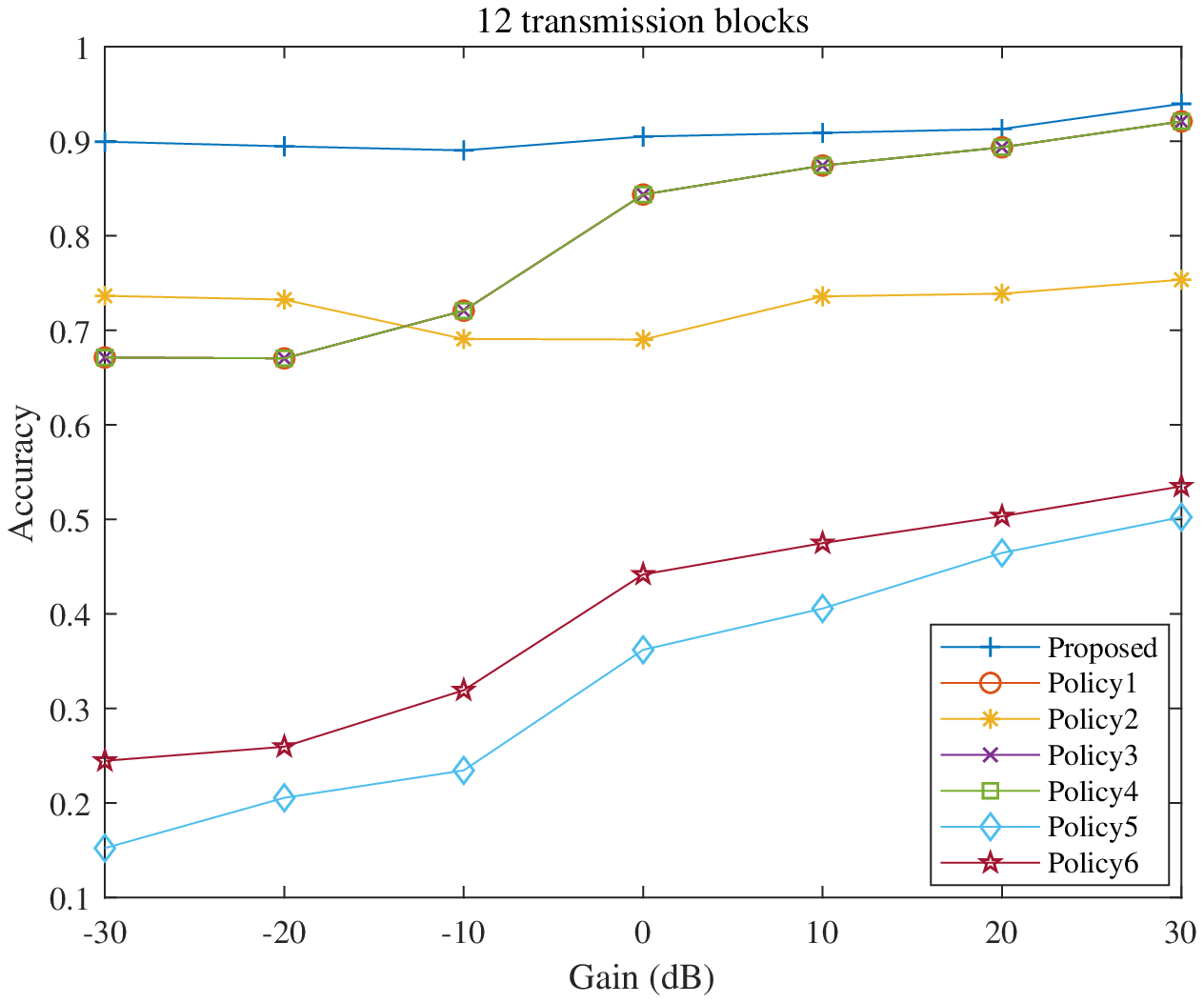}}
	\caption {Back-end accuracy of target model with reward function 2 compared to other policies.}
	\label{FIG6}
\end{figure*}

\begin{figure*}[!h]\label{fig7} 
	\centering
	\includegraphics[width=14cm]{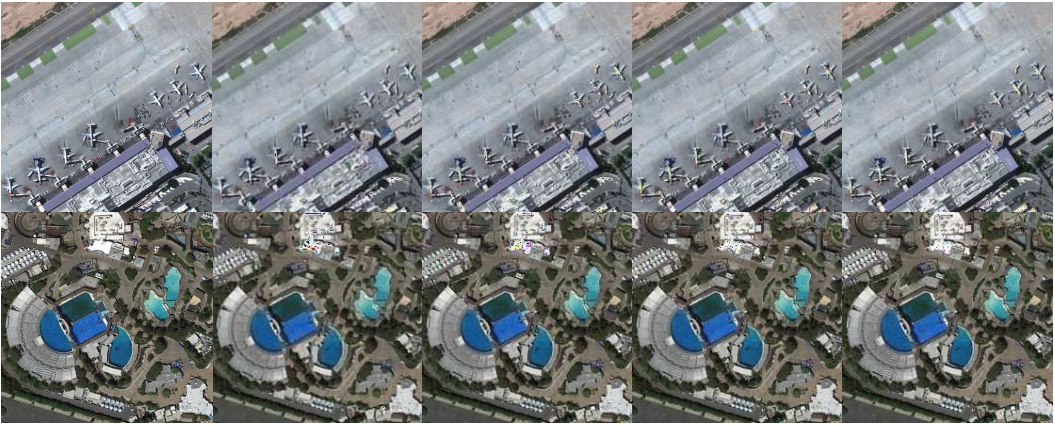}
	\caption {Deep reconstruction of full HR images at the final stage.}
	\label{FIG7}
\end{figure*}

\begin{figure*}[!h]\label{fig8} 
	\centering
	\subfigure[$\mathcal{R}_1=1/(1+2{T}),\eta_1=0.15$]{
		\label{fig8:subfig:a} %% label for first subfigure
		\includegraphics[width=14cm]{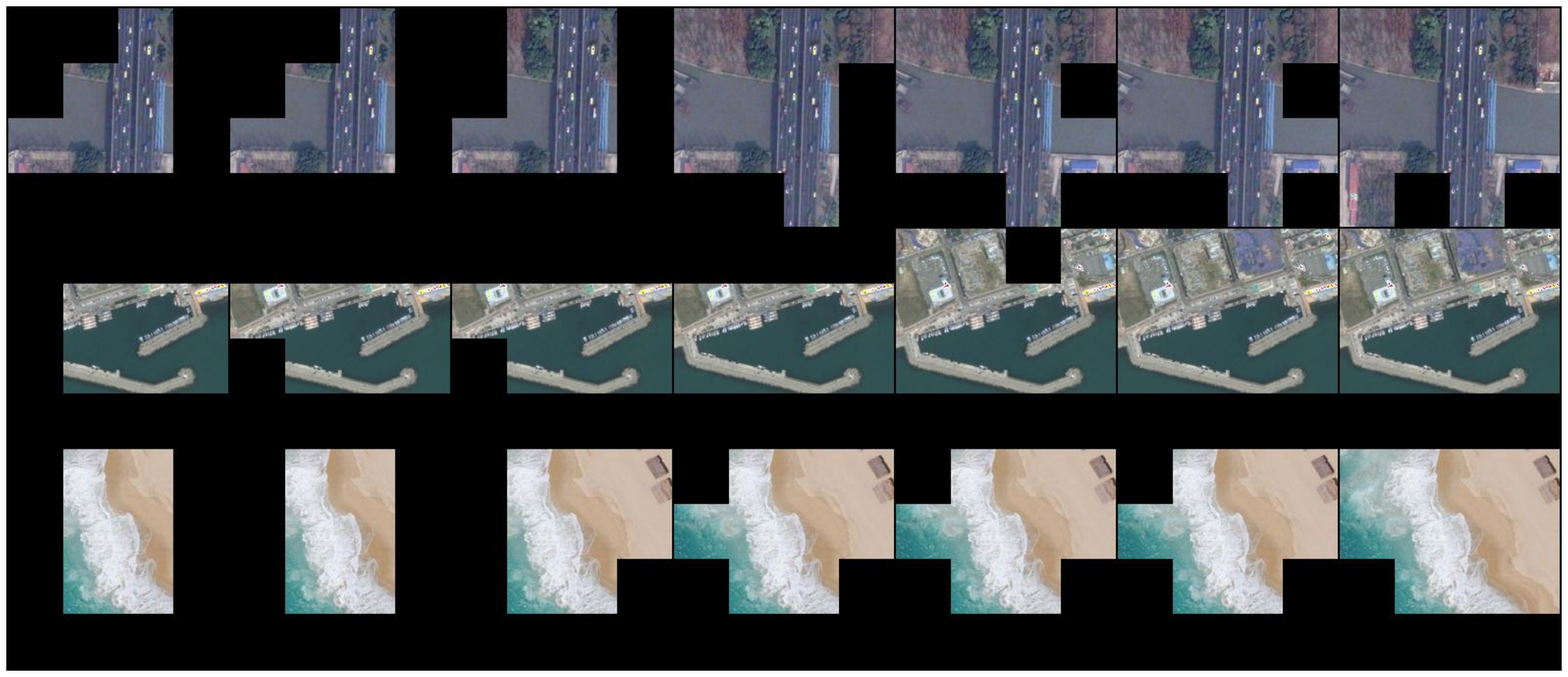}}
	\subfigure[$\mathcal{R}_2=\exp(-2T),\eta_2=-0.02$]{
		\label{fig8:subfig:b} %% label for second subfigure
		\includegraphics[width=14cm]{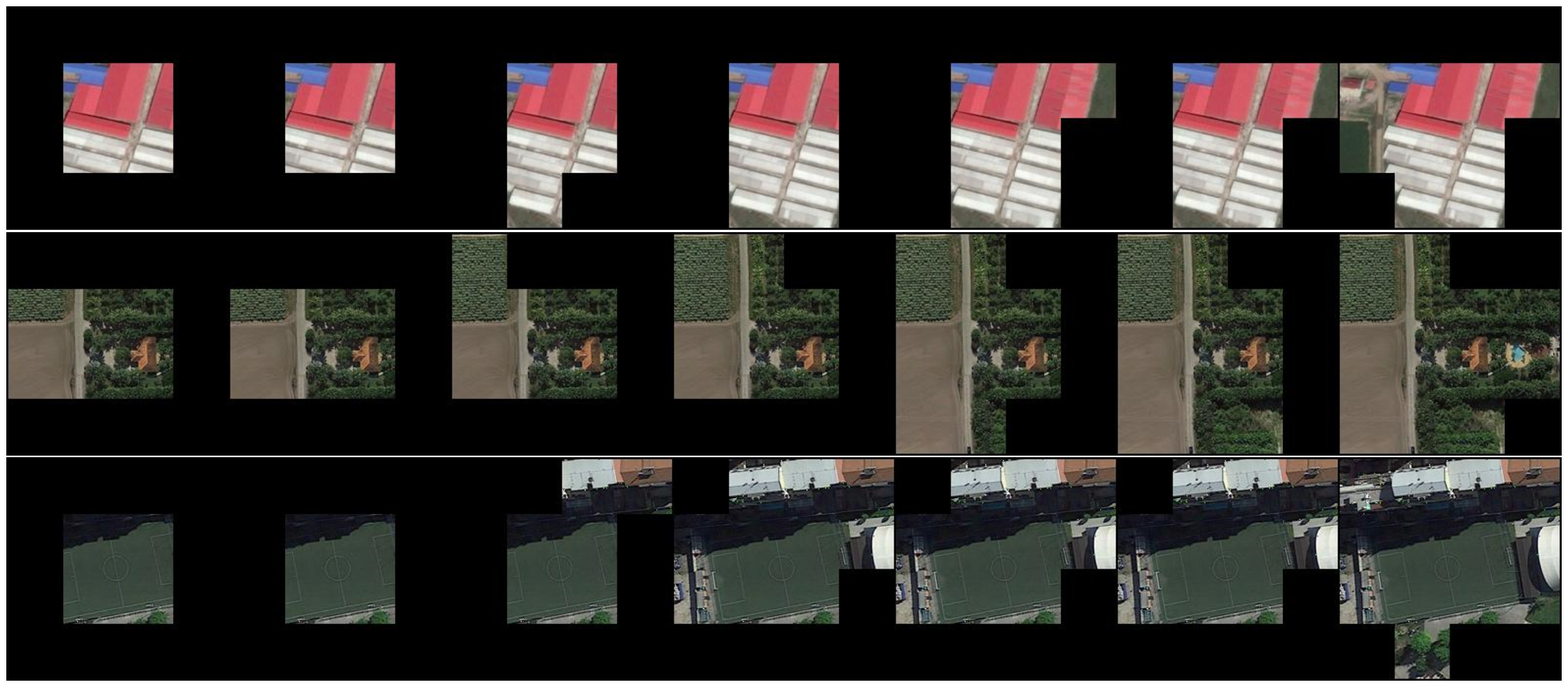}}
	\caption {Reconstruction results of semantic blocks with two reward functions.}
	\label{FIG8}
\end{figure*}

Now we will give a detailed introduction about several other comparative policies. $policy1$ selects semantic blocks from left to right in line order. $policy2$ extracts semantic blocks from top to bottom in the order of the columns. $policy3$ selects semantic blocks in a clockwise order from the center to the outside. On the contrary, $policy4$ extracts semantic blocks in a counterclockwise order from the center to the outside. $policy5$ randomly chooses $N_{i,j}$ semantic blocks that are the same as the proposed method for transmission each time. $policy6$ selects semantic blocks in order from largest to smallest in accordance with the saliency of semantic blocks. Tab. \ref{TAB1} reveals that the accuracy of proposed aerial image classification task offloading scheme is at least two percentage points higher than that of other traditional methods regardless of the channel conditions. Moreover, the proposed method can adaptively adjust the optimal policy of semantic extraction based on the change of channel conditions, focusing more on the semantic transmission latency when the channel conditions are poor while focusing on the completion of back-end classification task under better channel conditions.

Fig. \ref{FIG4} shows the number distribution of transmission blocks under different channel conditions with two different reward functions. The horizontal axis represents block numbers $N^a$ for transmission and the vertical axis represents number of samples that transmit $N^a$ semantic blocks ${Num}_j(N^a)=\sum_{i=1}^{N_{test}}1_{\{N_{i,j}=N^a\}}$, where $1_{\{\cdot\}}$ indicates a 0-1 function being 1 when the condition in brackets is met, 0 otherwise. It is pretty obvious that the model tends to transmit more semantic blocks under high channel gain and tends to transmit fewer semantic blocks under low channel gain.

The back-end accuracies of target model with two reward functions compared to other policies under different channel conditions are shown in Fig. \ref{FIG5} and Fig. \ref{FIG6}. The horizontal axis in each figure still shows 7 different channel gains, and the vertical axis denotes the classification accuracy of the back-end model under the constraints of $N^a$ transmission blocks. The number of samples successfully classified by the back-end model under the constraint of $N^a$ is ${Num}_j^\prime(N^a)\sum_{i=1}^{N_{test}}1_{\{N_{i,j}=N^a\wedge\mathop{\arg\max}\limits_{k}f_k^{tar}=C\}}$. Therefore, the accuracy on the curves can be calculated as ${Num}_j^\prime(N^a)/\sum_{m=1}^{N_{test}}1_{\{N_{i,j}=N^a\}}$. The two reward functions of proposed system can make the back-end accuracy better than other policies under the constraints of the same channel gain and transmission volume, especially in a poor channel environment.

Deep reconstruction of full HR images at the final stage are shown in Fig. \ref{FIG7}, where from left to right are the original image, reconstructed full image with $sr=\{0.1,0.2,0.3,0.4\}$. Fig. \ref{FIG8} shows the policies learned with two reward functions. From left to right, they are the semantic blocks reconstructed under 7 channel conditions.

\section{Conclusion}
In this article, we have studied the task oriented communication algorithm for image classification task offloading in aerial systems. In the scenario of aerial image transmission for scene classification task, a joint semantic extraction and compression model is proposed. Under the guidance of the policy gradient-based deep reinforcement learning algorithm, considering the system's uplink transmission latency and the classification accuracy of the back-end target model as the optimization objective, the proposed system can make optimal semantic extraction decisions under different channel conditions. Simulation results have shown that the proposed joint perception for decision scheme is feasible that a optimal balance between uplink transmission latency and classification accuracy can be achieved. Compared to the traditonal methods, it significantly improves the classification accuracy under the same transmission condition.

\bibliographystyle{IEEEtran}
\bibliography{IEEEabrv,xukang}

\end{document}